\documentclass[letterpaper, 10 pt, conference]{ieeeconf}  

\IEEEoverridecommandlockouts                              

\usepackage{graphics} 
\usepackage{graphicx}
\usepackage{epsfig} 
\usepackage{amsmath} 
\usepackage{amssymb}  
\usepackage{multirow}
\usepackage{subfigure}

\title{\LARGE \bf
Psi-Net: Shape and boundary aware joint multi-task deep network for medical image segmentation $\dagger$
}

\author{Balamurali Murugesan$^{\star}$, Kaushik Sarveswaran$^{\star}$, Sharath M Shankaranarayana, \\ 
Keerthi Ram, Jayaraj Joseph and Mohanasankar Sivaprakasam 
\thanks{$\star$ Contributed equally}
\thanks{$\dagger$ https://github.com/Bala93/Multi-task-deep-network/}
\thanks{Balamurali Murugesan and Mohanasankar Sivaprakasam are with Indian Institute of Technology Madras (IITM), India and Healthcare Technology Innovation Centre (HTIC), IITM, India (email: balamurali@htic.iitm.ac.in)}%
\thanks {Kaushik Sarveswaran is with Indian Institute of Information Technology Design \& Manufacturing Kancheepuram (IIITDM), India and HTIC, IITM, India}
\thanks {Keerthi Ram and Jayaraj Joseph are with HTIC, IITM, India}
\thanks{Sharath M Shankaranarayana is with Zasti, India}%
}

\begin{document}
\maketitle
\thispagestyle{empty}
\pagestyle{empty}
\begin{abstract}
Image segmentation is a primary task in many medical applications. Recently, many deep networks derived from U-Net has been extensively used in various medical image segmentation tasks. However, in most of the cases, networks similar to U-net produce coarse and non-smooth segmentations with lots of discontinuities. To improve and refine the performance of U-Net like networks, we propose the use of parallel decoders which along with performing the mask predictions also perform contour prediction and distance map estimation. The contour and distance map aid in ensuring smoothness in the segmentation predictions. To facilitate joint training of three tasks, we propose a novel architecture called Psi-Net with a single encoder and three parallel decoders (thus having a shape of $\Psi$), one decoder to learn the segmentation mask prediction and other two decoders to learn the auxiliary tasks of contour detection and distance map estimation. The learning of these auxiliary tasks helps in capturing the shape and the boundary information. We also propose a new joint loss function for the proposed architecture. The loss function consists of a weighted combination of Negative Log Likelihood and Mean Square Error loss. We have used two publicly available datasets: 1) Origa dataset for the task of optic cup and disc segmentation and 2) Endovis segment dataset for the task of polyp segmentation to evaluate our model. We have conducted extensive experiments using our network to show our model gives better results in terms of segmentation, boundary and shape metrics. 
\end{abstract}

\section{INTRODUCTION}
Image segmentation is the process of delineating structures of importance from an image. Identifying these structures in the medical image finds application in many medical procedures. To state some of them: 1) segmentation of optic cup and disc in the retinal fundus image is useful in glaucoma screening, 2) segmentation of polyp in colonoscopy image is helpful in cancer diagnosis, 3)  segmentation of the organ, bones benefit surgery planning and 4) segmentation of  lung  nodules  in  chest  Computed  Tomography  aids physicians to  differentiate malignant  lesions  from  benign  lesions. In recent years, deep learning networks \cite{survey} are widely used in medical image segmentation, and the most commonly used deep learning network is UNet \cite{unet}. 

\begin{table}[th]
\centering
\footnotesize
\caption{Summary of pros and cons of the models.}
\label{table:summary}
\begin{tabular}{|l|c|c|c|c|}
\hline
\multicolumn{1}{|c|}{} & \cite{unet} & \cite{dcan} & \cite{isbi_dcan} & Ours \\ \hline
Shape information & x & \checkmark & \checkmark & \checkmark \\ \hline
Class imbalance & x & x & \checkmark & \checkmark \\ \hline
Smooth boundary & x & x & \checkmark & \checkmark \\ \hline
Multiple object instances & \checkmark & \checkmark & x & \checkmark \\ \hline
\end{tabular}
\end{table}

UNet \cite{unet} is an encoder-decoder type of network which takes an image as input and outputs a pixel-wise classification probability score with cross-entropy as its loss function. This network has set new state of the art results for different medical image segmentation tasks. But there are some drawbacks with the architecture type, and loss functions used. For instance, the encoder block of the network undersamples the input through max-pooling layers which results in loss of spatial information. Similarly, having pixel-wise classification alone as a loss function produces uneven mask boundaries and outliers. In addition to this, the loss function doesn't take shape information into account which can help in performance improvement. Also, using cross-entropy as a loss function introduces class imbalance problem for images in which background dominates the object of interest which is very common in the medical images. To overcome the above-mentioned issues, multiple works have been reported in the literature \cite{survey}. 
In that, the architecture and loss functions followed by \cite{dcan} and \cite{isbi_dcan} are of our interest. Both these works use a similar architecture with a single encoder and two parallel decoders. The decoders are used for mask and contour prediction in \cite{dcan} whereas in \cite{isbi_dcan} it is used for estimating mask and distance map. Contour and distance map estimation act as regularizers to mask prediction. Shape information is imposed through contour and distance map in \cite{dcan} and \cite{isbi_dcan}. Class imbalance problem is mitigated in \cite{isbi_dcan} through its joint classification and regression approach while it will still be an issue in \cite{dcan} because of both decoders acting as classifiers. The boundaries obtained using \cite{isbi_dcan} are smooth and the segmentation has reduced outliers compared to \cite{unet,dcan}. But in multi-instance object segmentation cases, an object of smaller size can be treated as an outlier resulting in unsatisfactory segmentation. The summary of the above discussion are shown in Table \ref{table:summary}. 

\begin{figure*}[th]
  \centering
  \includegraphics[width=0.95\linewidth]{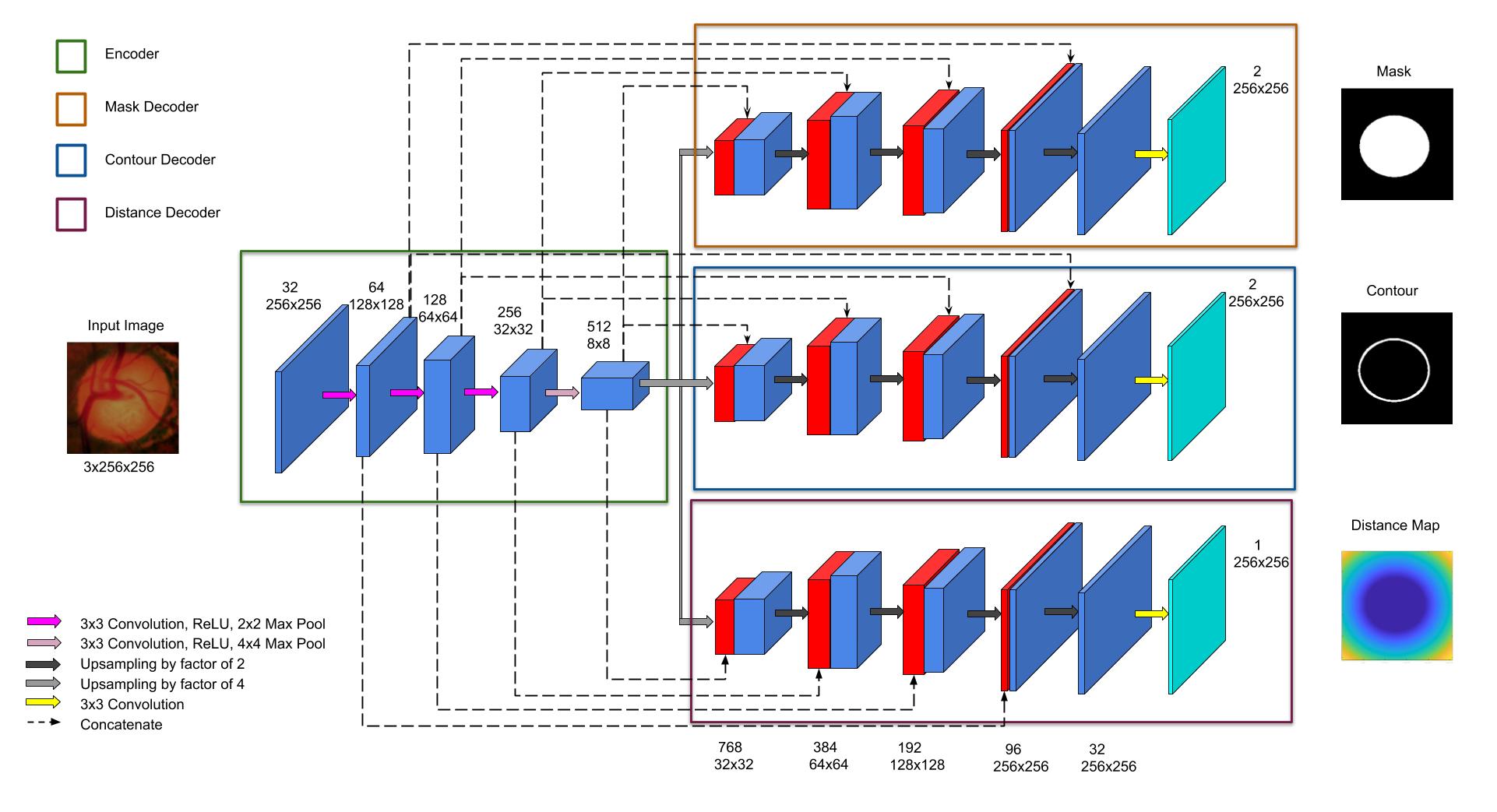}
  \caption{Psi-Net: Proposed architecture for segmentation with a single encoder and three decoders.}
  \label{fig:architecture}
\end{figure*}

The main contributions of our paper are as follows:

\begin{itemize}
    \item We propose a novel multi-task network Psi-Net with a single encoder and three decoders (architecture with shape $\Psi$). The decoders are used to learn three different tasks in parallel. The mask prediction is the primary task while the contour detection and distance map estimation are auxiliary tasks. These additional tasks are used to regularize the mask prediction path to produce a refined mask with smooth boundaries. 
    \item We propose a novel joint loss function to handle the three different tasks together. The joint loss function consists of a combination of Negative Log Likelihood (NLL) for mask, Negative Log Likelihood (NLL) for contour, and Mean Square Error (MSE) for distance.
    \item We qualitatively compared our results with \cite{unet}, \cite{dcan} and \cite{isbi_dcan}. The following evaluation metrics are used to perform a quantitative comparison: 
    \begin{itemize}
        \item \textit{Segmentation metrics} : Jaccard and Dice coefficients. 
        \item \textit{Shape similarity metrics} : Hausdorff distance
        \item \textit{Boundary metrics} : Segmentation evaluation around boundaries using trimap method.
    \end{itemize}
    The comparative study showed that our network performed better than others in all the evaluation metrics producing a better mask with smooth boundaries. 
\end{itemize}

\section{METHODOLOGY}

\subsection{Architecture}   
 
The architecture Psi-Net is a UNet-like encoder-decoder network, with one contracting encoder path on the left and three expansive structurally similar decoder paths on the right. The shape of the architecture resembles the mathematical symbol $\Psi$. The encoder path consists of repeated downsampling operations which halves the size of feature map at each stage. Each downsampling operation is preceded by a convolution operation with kernel size 3x3 and stride 1, which is followed by a Rectified Linear Unit (ReLU) activation. Each decoder block is symmetric to the encoder, and at each decoder layer, the features from the corresponding encoder layer are concatenated which helps in retaining multi-scale features. The final convolutional layer in the encoder is upsampled by a factor of 4 and given as input to the decoder blocks.

Each decoder block is trained for a different tasks - Mask segmentation, contour extraction and distance map estimation. The former two are pixel-wise classification tasks while the latter is a regression task. The blocks are identical in structure until the last layer, where a 3x3 convolution is applied, and the number of output channels is 1 in the distance decoder block and is equal to the number of input classes in the other two blocks. The outline of proposed network is shown in Fig. \ref{fig:architecture}.

\subsection{Loss Function}
The loss function consists of three components - Negative Log Likelihood (NLL) loss for mask and contour decoder blocks, and Mean Square Error (MSE) loss for the distance decoder block. Mask prediction is regularized by both contour and distance map predictions. The total loss is given by 

\begin{equation}
\mathcal{L}_{total} = \lambda_{1} \mathcal{L}_{mask} + \lambda_{2} \mathcal{L}_{contour} + \lambda_{3} \mathcal{L}_{distance}    
\end{equation}
where $\lambda_{1},\lambda_{2},\lambda_{3}$ are scaling factors. 

The individual losses are formulated below.

\subsubsection{Mask}
\begin{equation}\label{eq:l_mask}
\mathcal{L}_{mask} = \sum_{\boldsymbol{x} \, \epsilon\, \Omega}log\, p_{mask}(\boldsymbol{x};l_{mask}(\boldsymbol{x}))
\end{equation}

$\mathcal{L}_{mask}$ denotes the pixel-wise classification error. $\boldsymbol{x}$ is the pixel position in image space $\Omega$. $p_{mask}(\boldsymbol{x};l_{mask})$ denotes the predicted probability for true label $l_{mask}$ after softmax activation function.\\

\subsubsection{Contour}
\begin{equation}
\mathcal{L}_{contour} = \sum_{\boldsymbol{x}\, \epsilon\, \Omega}log\, p_{contour}(\boldsymbol{x};l_{contour}(\boldsymbol{x})) 
\end{equation}

$\mathcal{L}_{contour}$ denotes the pixel-wise classification error. $p_{contour}(\boldsymbol{x};l_{contour})$ denotes the predicted probability for true label $l_{contour}$ after softmax activation function.\\

\subsubsection{Distance}
\begin{equation}
\mathcal{L}_{distance} = \sum_{\boldsymbol{x}\, \epsilon\, \Omega} (\hat{D}(\boldsymbol{x}) - D(\boldsymbol{x}))^2 
\end{equation}

$\mathcal{L}_{distance}$ denotes the pixel-wise mean square error. $\hat{D}(\boldsymbol{x})$ is the estimated distance map after sigmoid activation function while $D(\boldsymbol{x})$ is the ground-truth distance map.

\begin{table*}[]
\centering
\caption{Comparison of Segmentation and Shape metrics.}
\label{table:results1}
\begin{tabular}{|l|c|c|c|c|c|c|c|c|c|}
\hline
\multirow{2}{*}{Architecture} & \multicolumn{3}{c|}{Cup} & \multicolumn{3}{c|}{Disc} & \multicolumn{3}{c|}{Polyp} \\ \cline{2-10} 
 & Dice & Jaccard & Hausdorff & Dice & Jaccard & Hausdorff & Dice & Jaccard & Hausdorff \\ \hline
1Enc 1Dec M \cite{unet} & 0.8655 & 0.7712 & 14.832 & 0.9586 & 0.9215 & 8.802 & 0.8125 & 0.7323 & 24.133 \\ \hline
1Enc 2Dec MC \cite{dcan} & 0.8715 & 0.7803 & 14.775 & 0.9646 & 0.9324 & 8.992 & 0.8151 & 0.7391 & 22.737 \\ \hline
1Enc 2Dec MD \cite{isbi_dcan} & 0.8723 & 0.7807 & 14.814 & \textbf{0.9665} & \textbf{0.9358} & 9.538 & 0.8283 & 0.7482 & 22.686 \\ \hline
1Enc 3Dec MCD (Ours) & \textbf{0.8745} & \textbf{0.7848} & \textbf{14.541} & \textbf{0.9665} & \textbf{0.9358} & \textbf{7.268} & \textbf{0.8462} & \textbf{0.7721} & \textbf{21.143} \\ \hline
\end{tabular}
\end{table*}

\section{EXPERIMENTS AND RESULTS}

\subsection{Dataset and Pre-processing}

\subsubsection{Dataset Description}
We validated our proposed segmentation approach for the following two applications:

\begin{enumerate}
    \item \textbf{Optic cup and disc segmentation} : We use ORIGA dataset \cite{origa} for the task of optic disc and cup segmentation. The dataset consists of $650$ color fundus image with ground truth segmentations for optic disc and cup. The color fundus images are of dimension 256 $\times$ 256. Ellipse fit is applied to output segmentation mask. 
    
    \item \textbf{Polyp segmentation} : We also use Polyp segmentation dataset from MICCAI 2018 Gastrointestinal Image ANalysis (GIANA) \cite{polyp_dataset}. The dataset consists of 912 images with ground truth masks. The dataset is split into 70\% for training and 30\% for testing. The images are center cropped and resized to 256 $\times$ 256. 
\end{enumerate}

\subsubsection{Preprocessing}

The dataset contains only segmentation mask. But for training our model, we need ground truth contour and distance map. The contour map is obtained by estimating the boundaries of connected components. These boundaries are subsequently dilated with a disk filter of radius 5. The distance map is obtained by applying an euclidean distance transform to the mask. The final distance map will contain zeros in the mask region, with the rest of the pixels denoting the shortest distance between that pixel and the mask boundary. 

\subsection{Implementation Details}
All the models are implemented using PyTorch. Models are trained for 150 epochs using Adam optimizer, with a learning rate of 1e-4 and batch size 4. Experiments have been conducted with NVIDIA GeForce GTX 1060 GPU - 6GB RAM.  

\subsection{Evaluation metrics}
In this section, A corresponds to the output of the method and B to the actual ground truth. 
\subsubsection{Segmentation evaluation}
Jaccard index and Dice similarity score are the most commonly used evaluation metrics for segmentation. Jaccard index (also known as intersection over union, IoU) is defined as the size of the intersection divided by the size of the union of the sample sets, and it is calculated as follows:
\begin{equation}
    Jaccard(A,B) = \frac{|A \cap B|}{|A \cup B|}
\end{equation}
\begin{equation}
    Dice(A,B) = \frac{2|A \cap B|}{|A| + |B|}
\end{equation}


\subsubsection{Shape Similarity}
The shape similarity is measured by using the Hausdorff distance between the shape of segmented object and that of the ground truth object, defined as
\begin{equation}
H(A,B) = max\Bigg\{ \underset{x \epsilon A}{sup} \; \underset{y \epsilon B}{ inf}\:||x-y||,\, \underset{y \epsilon B}{sup} \; \underset{x \epsilon A}{inf}\:||x-y|| \Bigg\}    
\end{equation}




\subsection{Results and Discussion}

Some of the abbreviations which will be used in this section are Encoder (Enc), Decoder (Dec), Mask (M), Contour (C) and Distance (D). The results of the proposed network (1Enc 3Dec MCD) is compared with the following networks. 

\begin{itemize}
    \item {A network (1Enc 1Dec M)  \cite{unet} with a single encoder and a decoder having NLL as loss function for mask prediction.}
    \item {A network (1Enc 2Dec MC) \cite{dcan} with a single encoder and two decoders having NLL as loss function for both mask and contour estimation.}
    \item {A network (1Enc 2Dec MD) \cite{isbi_dcan} with a single encoder and two decoders having NLL as loss function for mask and MSE as loss function for distance map estimation.}
\end{itemize}

\subsubsection{Standard Evaluation}
From Table \ref{table:results1} it can be seen that the network 1Enc 3Dec MCD has shown better performance in Dice and Jaccard compared to the networks 1Enc 1Dec M, 1Enc 2Dec MC and 1Enc 2Dec MD. This improvement in performance can be attributed to the use of two auxiliary regularizers, in the form of contour detection and distance map estimation, as opposed to a single regularizer in 1Enc 2Dec MC and 1Enc 2Dec MD. Both the networks 1Enc 2Dec MC and 1Enc 2Dec MD use shape information for mask refinement. While 1Enc 2Dec MD provides smooth boundaries compared to  1Enc 2Dec MC, it has a drawback in handling multiple object instances which is not an issue in 1Enc 2Dec MC. Since both these networks complement one another, combining these models brings the best result. The segmentation of polyp is relatively difficult when compared to optic cup and disc segmentation because of its large variations in size and shape. From Table  \ref{table:results1}, it is evident that our network shows substantial improvement in performance for polyp segmentation compared to optic cup and disc segmentation. 

\subsubsection{Shape Similarity}
Along with better segmentation, the network should also produce segmentation maps which are similar to ground truth masks regarding shape \cite{dcan}. This shape similarity is obtained by Hausdorff distance. From Table \ref{table:results1}, it is clear that our network does well in capturing shape information compared to other networks. Also, sorting the Hausdorff distance helps in coming to the following inferences: 1) the addition of auxiliary tasks does help in preserving shape. 2) the auxiliary task of distance map estimation captures the shape better than the contour extraction. 

\subsubsection{Segmentation around boundaries}
\begin{figure}[]
  \centering
  \includegraphics[width=0.85\linewidth]{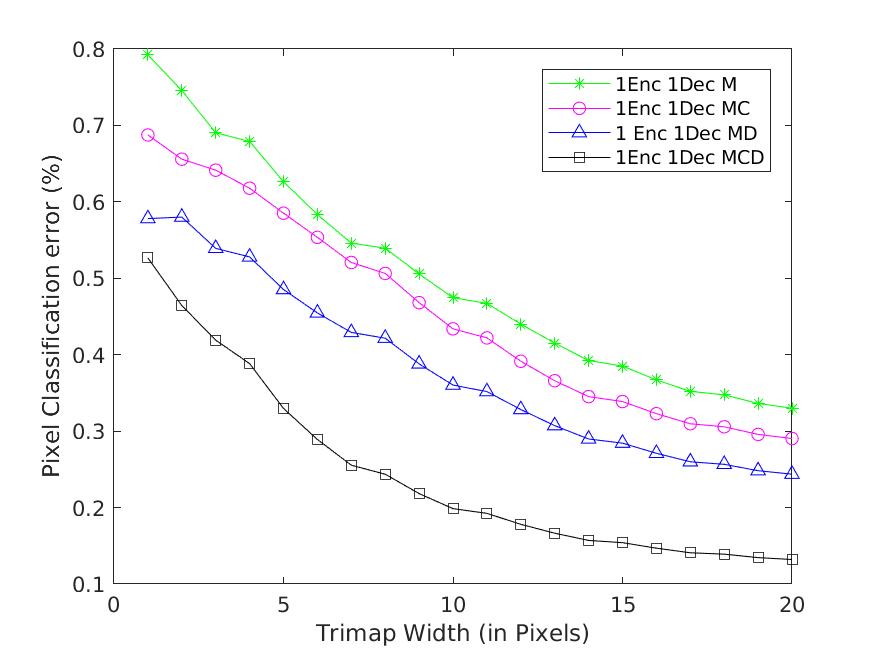}
  \caption{Percent of misclassified pixels within trimaps of different widths.}
  \label{fig:plot}
\end{figure}

In the above paragraphs, we have mentioned that our network produces segmentation masks with smooth boundaries. Smooth boundaries indicate a better segmentation around the boundary. We evaluated the segmentation accuracy around boundary with the method adopted in \cite{nips_trimap}. Specifically, we count the relative number of misclassified pixels within a narrow band (“trimap”) surrounding actual object boundaries, obtained from the accurate ground truth images. As can be seen in Figure \ref{fig:plot}, our method has less error for trimaps of different widths.

\begin{figure}
\centering
\subfigure{\includegraphics[width=12mm]{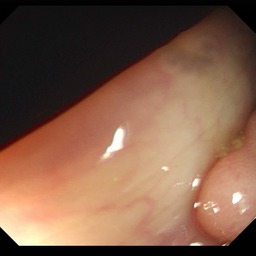}}
\subfigure{\includegraphics[width=12mm]{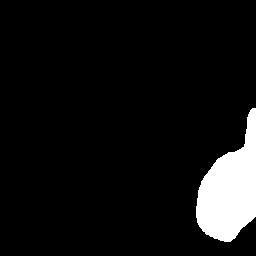}}
\subfigure{\includegraphics[width=12mm]{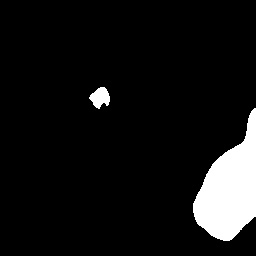}}
\subfigure{\includegraphics[width=12mm]{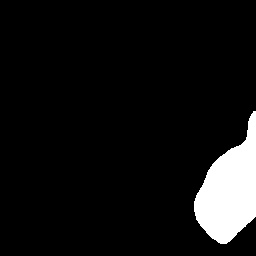}}
\subfigure{\includegraphics[width=12mm]{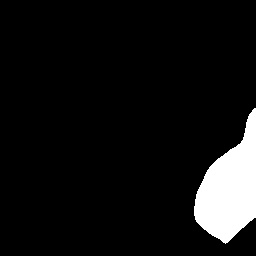}}
\subfigure{\includegraphics[width=12mm]{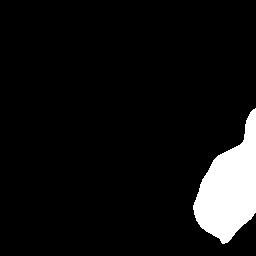}}

\subfigure{\includegraphics[width=12mm]{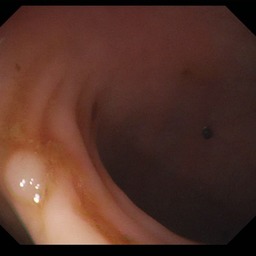}}
\subfigure{\includegraphics[width=12mm]{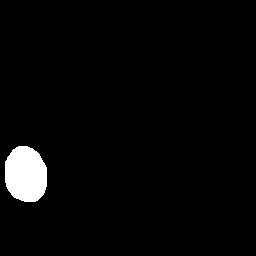}}
\subfigure{\includegraphics[width=12mm]{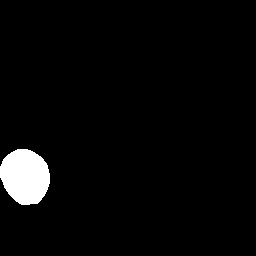}}
\subfigure{\includegraphics[width=12mm]{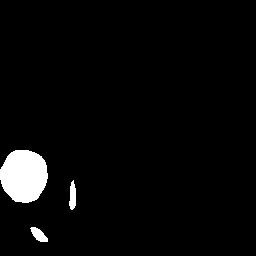}}
\subfigure{\includegraphics[width=12mm]{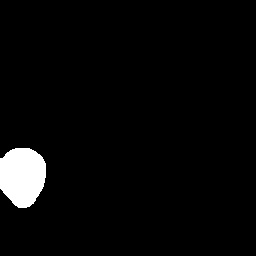}}
\subfigure{\includegraphics[width=12mm]{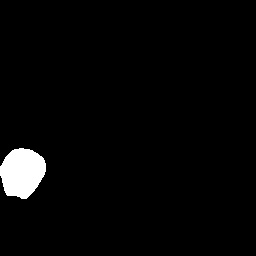}}

\subfigure{\includegraphics[width=12mm]{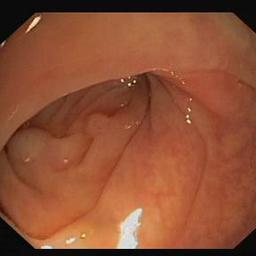}}
\subfigure{\includegraphics[width=12mm]{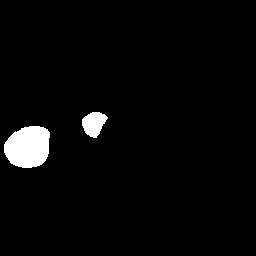}}
\subfigure{\includegraphics[width=12mm]{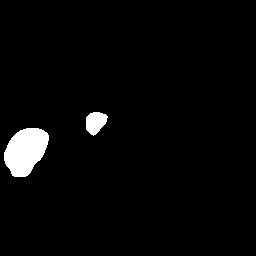}}
\subfigure{\includegraphics[width=12mm]{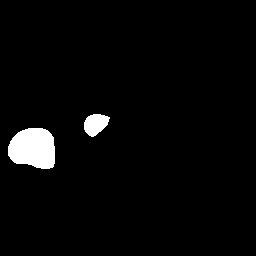}}
\subfigure{\includegraphics[width=12mm]{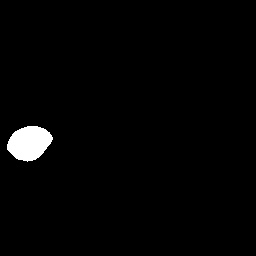}}
\subfigure{\includegraphics[width=12mm]{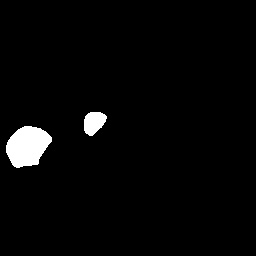}}

\subfigure{\includegraphics[width=12mm]{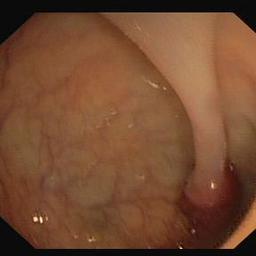}}
\subfigure{\includegraphics[width=12mm]{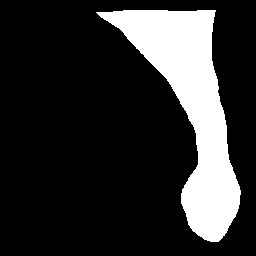}}
\subfigure{\includegraphics[width=12mm]{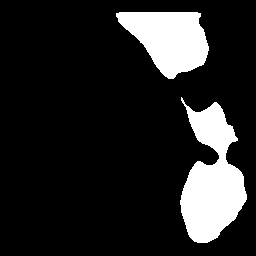}}
\subfigure{\includegraphics[width=12mm]{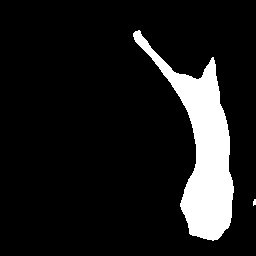}}
\subfigure{\includegraphics[width=12mm]{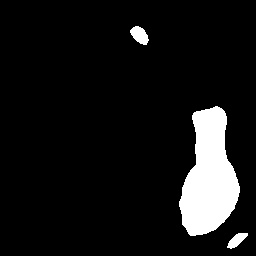}}
\subfigure{\includegraphics[width=12mm]{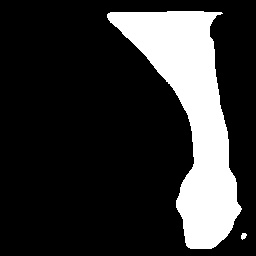}}

\caption{From left to right : Image, Ground truth mask, Predicted mask of \cite{unet,dcan,isbi_dcan} and Ours.}
\label{fig:results4}
\end{figure}

\subsubsection{Qualitative comparison}
The qualitative comparison of our network 1Enc 3Dec MCD with 1Enc 1Dec M, 1Enc 2Dec MC and 1Enc 2Dec MD can be seen in Fig. \ref{fig:results4}. To better appreciate the improvement of our model we have shown only the polyp dataset outputs. The mask predicted by our network and 1Enc 2Dec MD is smooth without outliers compared to the mask predicted by the networks 1Enc 1Dec M and 1Enc 2Dec MC. This is depicted in the first two rows of the figure. In the third row of the figure, it can be seen that the network 1Enc 1Dec MD fails in case of multi-instance object segmentation while our network performs well as that of 1Enc 2Dec MC. The fourth row shows a case where our network outperforms the other networks.




\section{Conclusion}
In this paper, we have introduced a network called Psi-Net with a single encoder and three parallel decoders. The three decoders are used for mask prediction, contour extraction and distance map estimation respectively. We have also introduced a joint loss function to optimize the proposed network. We have shown that this kind of architecture preserves shape well with better boundary outputs and improved segmentation performance. 

\bibliographystyle{IEEEtran}
\bibliography{embc}

\end{document}